\documentclass{article} 
\usepackage[preprint]{colm2026_conference}

\usepackage{microtype}
\usepackage{hyperref}
\usepackage{url}
\usepackage{booktabs}

\usepackage{lineno}

\definecolor{darkblue}{rgb}{0, 0, 0.5}
\hypersetup{colorlinks=true, citecolor=darkblue, linkcolor=darkblue, urlcolor=darkblue}

\usepackage{multirow}
\usepackage{colortbl}
\usepackage{xspace}

\newcommand{\ASYNCIO}{\textit{Asynchronous I/O}\xspace}
\newcommand{\SPECTOOL}{\textit{Speculative Tool Calling}\xspace}

\newcommand{\sig}[1]{%
  \ifnum#1=1 $^{*}$%
  \else\ifnum#1=2 $^{**}$%
  \else\ifnum#1=3 $^{***}$%
  \else%
  \fi\fi\fi%
}

\definecolor{darkblue}{rgb}{0, 0, 0.5}
\hypersetup{colorlinks=true, citecolor=darkblue, linkcolor=darkblue, urlcolor=darkblue}

\definecolor{berkeleyblue}{HTML}{3B7EA1}
\definecolor{berkeleygold}{HTML}{FDB515}
\definecolor{customgray}{HTML}{888888}
\definecolor{darkgray}{HTML}{222222}
\definecolor{main}{HTML}{4472C4}   
\definecolor{sub}{HTML}{EBF4FF} 
\definecolor{github}{HTML}{181717}
\definecolor{hf}{HTML}{FC9313}

\definecolor{berkeleyblue}{HTML}{3B7EA1}
\definecolor{berkeleygold}{HTML}{FDB515}
\definecolor{customgray}{HTML}{888888}
\definecolor{darkgray}{HTML}{222222}
\definecolor{main}{HTML}{4472C4}   
\definecolor{sub}{HTML}{EBF4FF}

\newcommand\hc{ \rowcolor{teal!10}}
\newcommand\hd{ \rowcolor{teal!18}}

\usepackage{caption}
\usepackage{quoting}
\usepackage{inconsolata}
\usepackage{url}
\usepackage{graphicx}
\usepackage{xspace}
\usepackage[subtle]{savetrees}
\usepackage[most]{tcolorbox}
\usepackage{colortbl}
\usepackage{microtype}      
\usepackage{xcolor}
\usepackage{multirow}
\usepackage{multicol}
\usepackage{longtable}
\usepackage{listings}
\usepackage{enumitem}
\usepackage{psfrag}
\usepackage{verbatim}
\usepackage{pifont}%
\usepackage{cleveref}
\usepackage{adjustbox}

\title{Speculative Interaction Agents: Building Real-Time Agents with Asynchronous I/O and Speculative Tool Calling}

\author{%
  Coleman Hooper$^{1*}$\enskip\enskip
  Minwoo Kang$^{1*}$\enskip\enskip
  Suhong Moon$^{1*}$\enskip\enskip
  Nicholas Lee$^{1}$\enskip\enskip
  Eric Wen$^{1}$\\[0.75mm]
  \textbf{
  John Wawrzynek$^{1}$\enskip\enskip
  Michael W. Mahoney$^{1,2,3}$\enskip\enskip
  Yakun Sophia Shao$^{1}$}\\[0.75mm]
  \textbf{
  Amir Gholami$^{1,2}$\enskip\enskip
  Kurt Keutzer$^{1}$}\\[0.75mm]
  $^{1}$University of California, Berkeley \enspace
  $^{2}$ICSI \enspace
  $^{3}$LBNL \\[0.75mm]
  \texttt{\small{Correspondence to: chooper@berkeley.edu}}\\
  \vspace{-2mm}\\
  \small{$^{*}$Equal contribution.}
}

\begin{document}

\ifcolmsubmission
\linenumbers
\fi

\maketitle

\begin{abstract}

There is a growing demand for agentic AI technologies for a range of downstream applications like customer service and personal assistants.
For applications where the agent needs to interact with a person, real-time low-latency responsiveness is required;
for example, with voice-controlled applications, under 1 second of latency is typically required for the interaction to feel seamless.
However, if we want the LLM to reason and execute an agentic workflow with tool calling, this can add several seconds or more of latency, which is prohibitive for real-time latency-sensitive applications.
In our work, we propose \textbf{Speculative Interaction Agents} to enable real-time interaction even for agents with complex multi-turn tool calling. 
We propose \textit{Asynchronous I/O}, which decouples the core agent reason-and-act thread from waiting for additional information from either the user or environment, thereby allowing for overlapping agentic processing while waiting on external delays.
We also propose \textit{Speculative Tool Calling} as a method to manage task execution when the agent is still unsure if it has received the full information or if additional user information may later be provided.
For strong cloud models, our method can be applied out-of-the-box to existing real-time cloud APIs, providing 1.3-1.7$\times$ speedups with minor accuracy loss.
To enable real-time interaction with small edge-scale models, we also present a clock-based training methodology that adapts the model to handle streaming inputs and asynchronous responses, and demonstrate a synthetic data generation strategy for SFT.
Altogether, this approach provides \textbf{1.6-2.2}$\times$ speedups with the Qwen2.5-3B-Instruct and Llama-3.2-3B-Instruct models across multiple tool calling benchmarks. 

\end{abstract}

\section{Introduction}

\begin{figure*}[t]
  \centering
  \includegraphics[width=\textwidth]{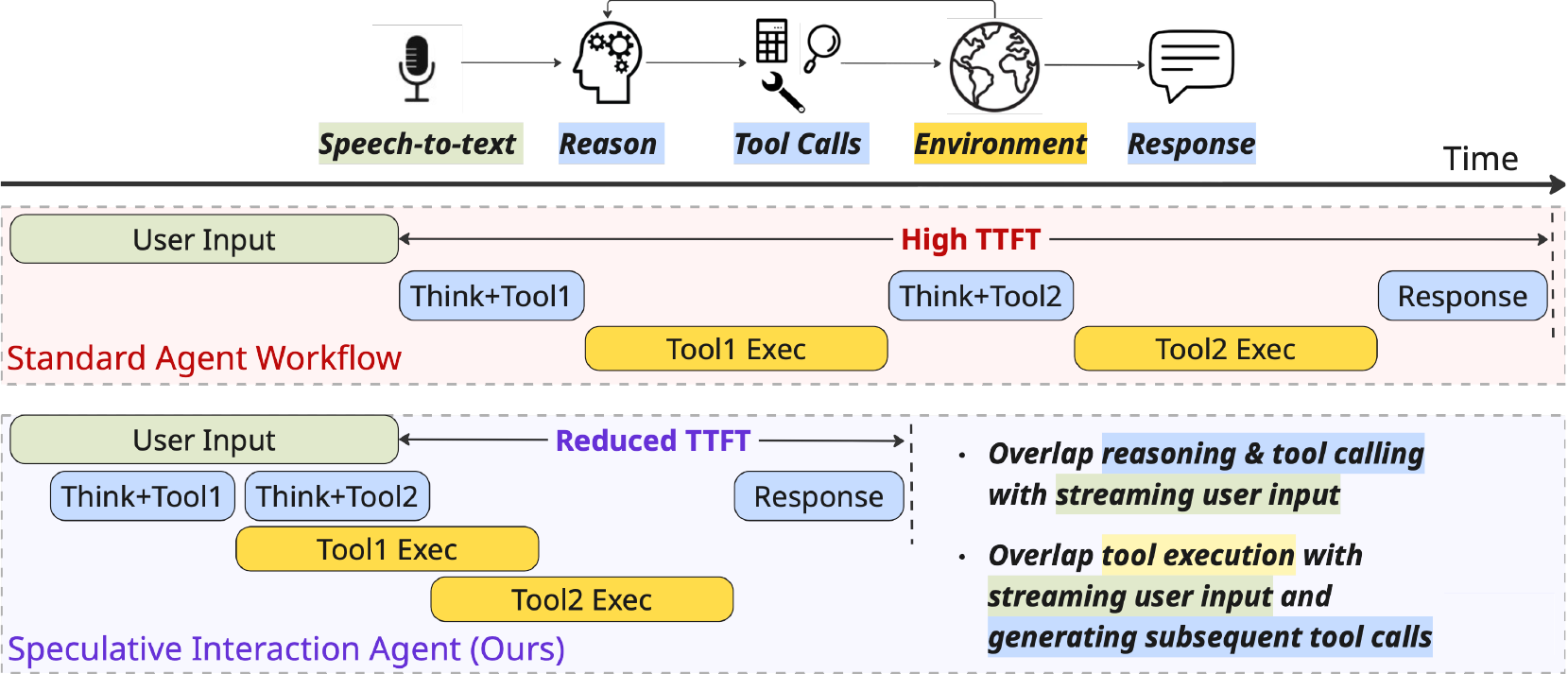}
  \caption{
  \textbf{Overview of Speculative Interaction Agent workflow.} A diagram outlining our proposed workflow in comparison to a standard agentic workflow.
  Standard agentic workflows have high time-to-first-token (TTFT) due to having to wait for the full user response before beginning to think and act on it, as well as having to pause to wait on tool execution.
  Our approach reduces latency by i) overlapping reasoning and acting with the streaming user input (making our agent asynchronous with respect to user inputs), and ii) overlapping reasoning and acting with tool execution (thereby making our agent asynchronous with respect to the environment).
  }
  \label{fig:intro-figure}
  \vspace{-3mm}
\end{figure*}

There is an increasing push for applications that use LLMs as agents, where the model thinks and then executes actions to interact with the external environment \citep{yao2022react,patil2024gorilla,schick2023toolformer}.
This enables the LLM to actuate changes in the real world, for example by looking up information in a database or sending a message.
There has also been growing interest in reasoning models which produce chain-of-thought before answering a question \citep{wei2022chain,guo2025deepseek,qwq32b}. 
Having the model generate this step-by-step chain enhances model accuracy.
However, the increased generation length where the model generates its reasoning process introduces additional latency.

Additionally, LLM agents are increasingly leveraged in real-time applications such as customer support and personal assistants.
For many real-time applications, users prefer voice as a medium of interaction to textual inputs via keyboard. 
For voice interactions to feel natural, less than 1 second of latency is typically required \citep{miller1968response}. 
However, if we want the LLM to do any complex agentic workflow which requires reasoning and calling tools, this adds several seconds of latency, which is prohibitive for real-time applications.
This latency comes from both long tool latencies as well as time for the model to reason before each action.

Previous methods have aimed to accelerate tool calling either through parallelizing tool calls \citep{kim2024llm} or by allowing models to reason while waiting on previous tool calls to execute \citep{gim2024asynchronous}.
Additionally, previous voice agent studies have proposed methods to enable the agent to begin execution before receiving the full input \citep{arora2025stream,chiang2025shanks}.
However, there is no unifying work that allows the model to reason and act while waiting on information from both the user and from the environment.
Additionally, existing methods for voice agents which begin processing based on partial user inputs may launch incorrect tool calls, and existing frameworks lack a method to correct these mistakes while preserving model accuracy.

In our work, we propose \textit{Speculative Interaction Agents}, which enable real-time interaction by decoupling the agent from the user and environment streams and allowing it to reason and act continuously.
We present our approach assuming streaming input text from speech-to-text using voice as a driving application, but our asynchronous approach is generalizable to any agentic application requiring real-time user interaction.
In particular, our work makes the following contributions:

\vspace{-3mm}
\begin{itemize}
\setlength\itemsep{2pt}
    \item We propose \ASYNCIO, where we decouple the agent's thinking and acting from delays with information arriving from the user and environment (see Section \ref{sec:asynchronous-io}). This decoupling allows the model to work continuously while awaiting updates.
    \item To handle cases where the user has not yet provided complete information, we propose \SPECTOOL, which allows for launching tool calls speculatively and later correcting any mistakes, and which also holds up sensitive tools from execution until they receive final confirmation (see Section \ref{sec:spec-tool-calling}).
    \item We present a clock-based training methodology for training agents to reason and act while waiting for information. We describe how we synthesize trajectories for training models using SFT (see Section \ref{sec:training}).
    \item We also present an inference-time framework which contains a core event-driven agent loop to manage information being received asynchronously from the user and environment, as well as a decoupled task execution module to manage in-flight tool calls (see Section \ref{sec:inference-framework}).
    With our framework, we observe 1.3-1.7$\times$ speedups for cloud real-time APIs and 1.6-2.2$\times$ speedups for edge-scale finetuned models, demonstrating how our approach enables real-time agentic interaction even with agents that have to perform complex multi-turn workflows.
\end{itemize}

\begin{figure*}[t]
  \centering
  \includegraphics[width=\textwidth]{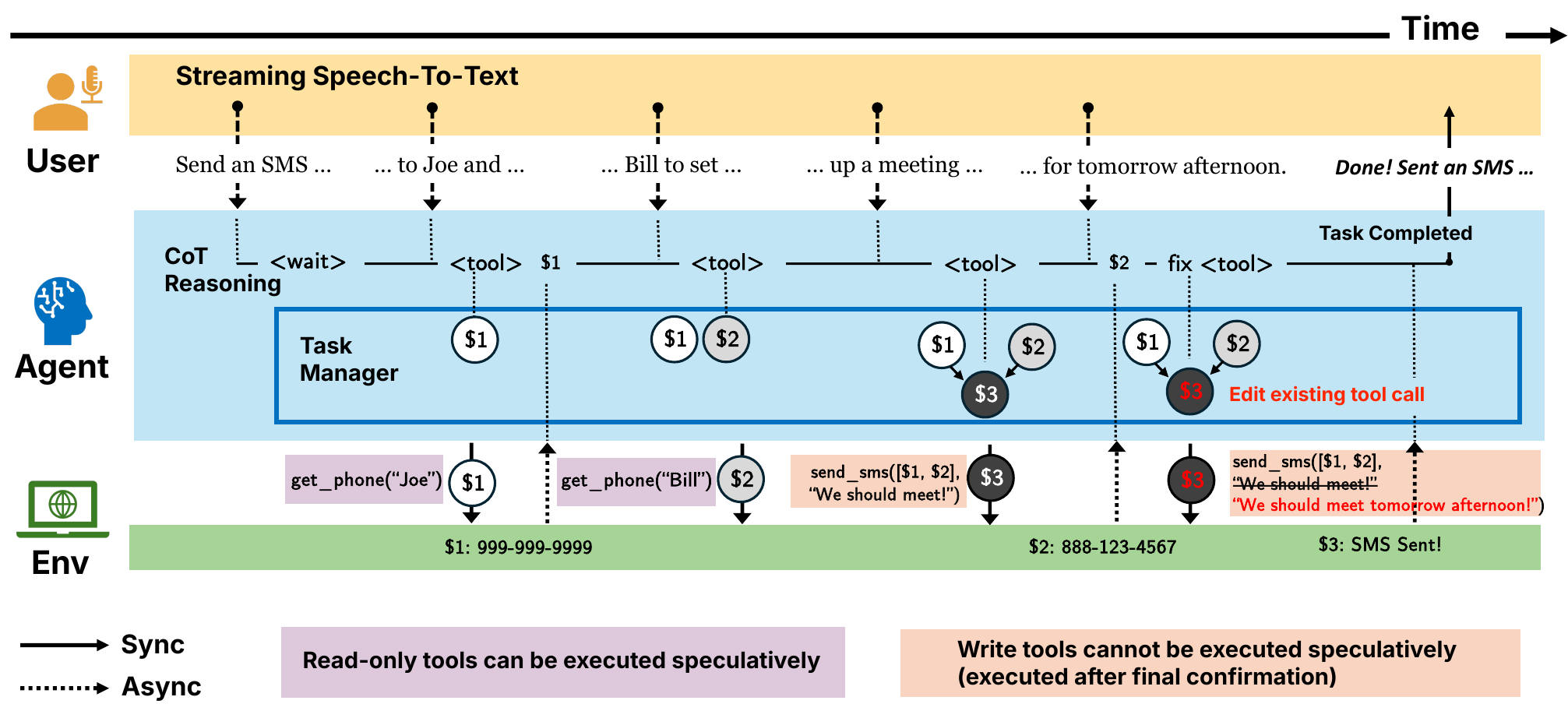}
  \caption{ 
  \textbf{Example Execution Trajectory.} An example trajectory showing agent execution unrolled over time, demonstrating how the agent reasons and acts continuously while waiting on additional user information as well as tool execution. 
  The diagram also highlights how tool calls can be generated based on partial user inputs, and how tool calls can later be modified after receiving more information to ensure the updated tool calls are aligned with the user query.
  Note that only read-only tools can be executed before receiving a final confirmation signal to avoid irreversibly modifying system state.
  }
  \label{fig:execution-trace}
\end{figure*}

\section{Related Work}

Here, we provide a brief summary of related work on efficient language and voice agents as well as related system infrastructure projects.
A more detailed description of related work and comparison with our method is provided in Appendix \ref{sec:appendix_related_work}.

\subsection{Efficient LLM Agents}

Several prior works have explored how to execute agentic workflows more efficiently, including parallelizing tool calls where possible \citep{kim2024llm} and performing work while waiting on asynchronous tool execution responses \citep{gim2024asynchronous}.
Another prior work used separate thinker and talker threads so that one thread could talk to the user while the other thread performed a background reasoning process \citep{yakushev2025asynchronous}.
An additional concurrent work trained an ``Interaction Model" to respond in real-time to the user, while delegating complex reasoning to a background thread \cite{thinkingmachines2026interactionmodels}.
There has also been work on designing proactive and interactive LLM agents, including Pepper \citep{rllm2025pepper}, which proposed an event-driven architecture for proactive LLM assistants.
Compared with prior work, our work builds on the asynchronous tool calling method from \citep{gim2024asynchronous} and generalizes this to handle both asynchronous user updates as well as asynchronous responses from the environment.
Our work also builds on the decoupled planning/execution from LLMCompiler \citep{kim2024llm}, which allows for tool calls to be executed in parallel as soon as their operands are ready.
We extend this to allow for issuing tool calls iteratively as we obtain sufficient information, and also propose \textit{Speculative Tool Calling} (Sec. \ref{sec:spec-tool-calling}) to hold up sensitive tool calls until they are confirmed and to allow for modifying/removing tool calls that have been issued but not yet finished execution.

\subsection{Efficient Voice Agents}

There has also been prior works that have explored methods to accelerate voice agents by exploiting the streaming nature of speech.
StreamRAG \citep{arora2025stream} proposes launching RAG calls partway through receiving the streaming user input, thereby reducing response latency.
SHANKS \citep{chiang2025shanks} processes the streaming input speech in fixed chunks and interleaves unspoken reasoning (as well as tool calls) between chunks while waiting for further information from the user, whereas STITCH \citep{chiang2025stitch} interleaves text-to-speech with unspoken reasoning.
Our approach builds on \citep{chiang2025shanks} by pursuing a similar approach where we interleave reasoning and acting with streaming input speech. 
The key differences with our method are 1) we support asynchronous tool calling as well as asynchronous user inputs, 2) we propose \textit{Speculative Tool Calling} (Sec. \ref{sec:spec-tool-calling}) to handle sensitive tool calls and cases where the tool call is issued incorrectly with partial information. 

\subsection{System Support for Streaming Inference}

There has also been recent work on system-level optimizations to enable real-time interactive agentic systems.
The OpenAI Realtime API \citep{openai_realtime_api} and Gemini Live API \citep{GoogleLiveAPI2025} both provide websocket-based interfaces that support streaming inputs (for reducing TTFT due to prefilling latency) and outputs (producing partial responses that can start being returned to the user before the full output is available).
Additionally, open-source serving frameworks like vLLM have recently added support for efficient streaming input processing \citep{vllm2026streaming}.
There has also been efforts to design infrastructure for efficient streaming processing at the edge \citep{jeffries2024moonshine}.

\begin{figure*}[t]
  \centering
  \includegraphics[width=0.9\textwidth]{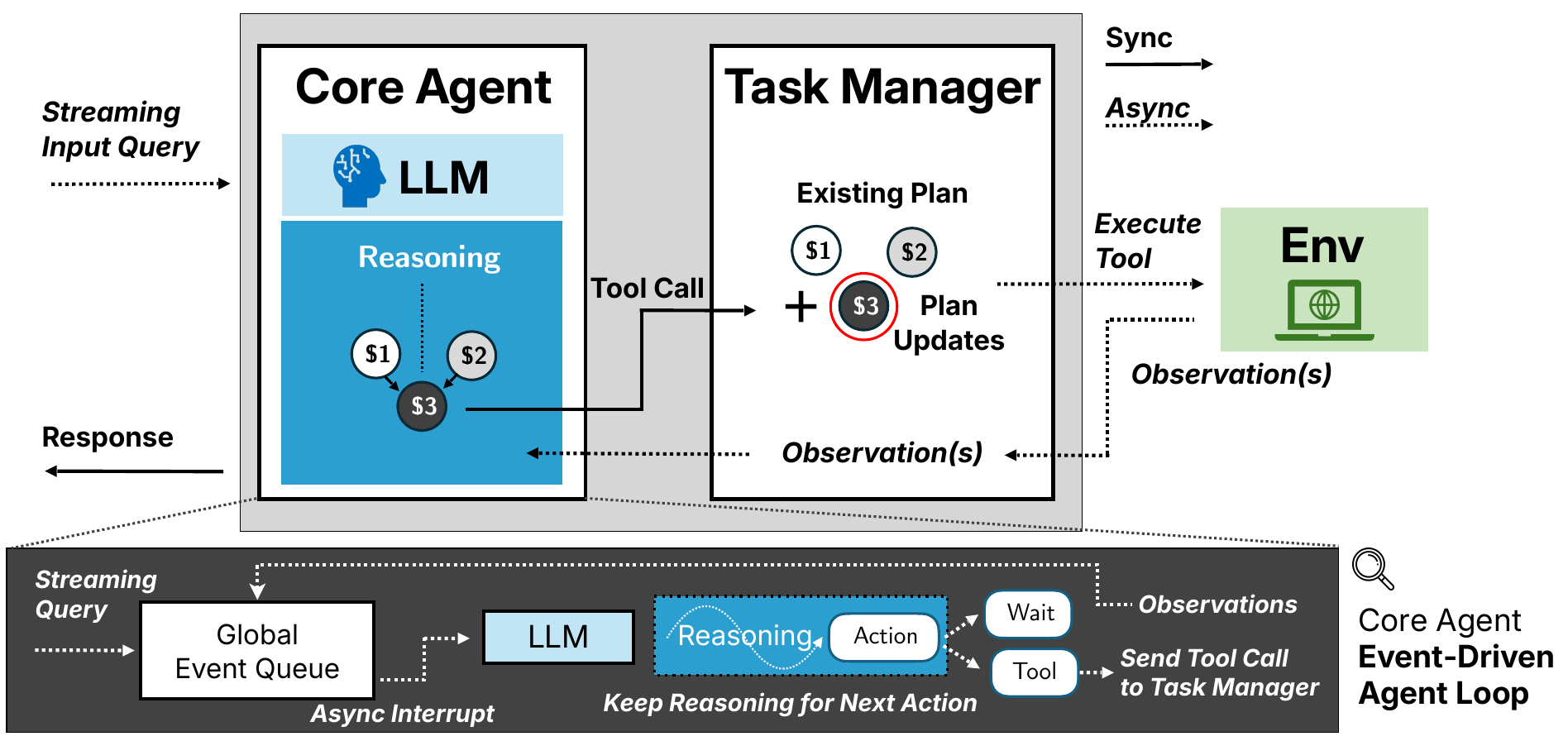}
  \caption{
  \textbf{Top:} Architecture diagram for our system implementation, which consists of: i) a core agent loop that receives streaming user inputs and tool responses from the environment and which launches tool calls as needed; and ii) a task manager that manages in-flight tool calls and tracks dependencies.
\textbf{Bottom:} A zoomed-in diagram showing the core event-driven control loop, which injects new information from the user and environment into the model's context as it arrives.
The model is allowed to reason and act continuously while waiting for additional information. 
  }
  \label{fig:system-architecture}
\end{figure*}

\section{Method}

\subsection{Asynchronous I/O}
\label{sec:asynchronous-io}

The aim of our methodology is to enable models to reason and act continuously while waiting on information from either the user or the environment. 
Building on prior work on LLM agent processing based on partial inputs for voice agents ~\citep{arora2025stream,chiang2025shanks,shih2025can}, as well as prior work on asynchronous tool use ~\citep{gim2024asynchronous}, we propose \textit{Asynchronous Input/Output (I/O)}, where we decouple the agent's thinking and acting stream from information coming from the user as well as information arriving from tool execution. 
This decoupling allows the model to work continuously while waiting for additional information from the user or from the environment.

We assume that the user input is provided incrementally as it is available, with updates provided in ``$<partial\_query\_update>$'' / ``$</partial\_query\_update>$'' tags (and with ``$<final\_query\_update>$'' / ``$</final\_query\_update>$'' used to denote the end of utterance from the user).
To enable Asynchronous I/O, we design a set of possible actions to enable the model to think and act asynchronously with respect to the user and environment.
At each step, the model must first produce reasoning (encompassed in ``$<think>$'' / ``$</think>$'' tags), and then produce one of the following actions:
\begin{enumerate}
    \item ``$<tool\_call> ID. tool\_name(a1, a2, ...)</tool\_call>$'' to call tool $toolname$ with arguments $a1, a2, ...$ (where $ID.$ is an integer starting from 1 that denotes the tool call ID, and which is used to label the tool call observation, which is returned asynchronously as ``$<information> ID. ...</information>$'')
    \item ``$<pause>$'' to halt generation and wait for more information from the user or environment if there is no other action required based on currently available information
    \item ``$<answer> ... </answer>$'' to produce the final answer
\end{enumerate}

The tool call IDs are used for tracking and editing tool calls before they are finalized (see Sec. \ref{sec:spec-tool-calling}).
When additional information from the user or environment becomes available while the model is generating, we halt the model's generation and inject the updated information (with ``$</think\_interrupted>$'' prepended to the information).

Additionally, there are forbidden cases which we must consider: i) the model should not pause for information if there are no outstanding tool calls and it has already received the final user update, ii) the model should not produce the final answer before receiving the full user query, and iii) cases where there are formatting errors with the generation or action.
For all these cases, instead of processing the invalid action, we inject an error message instructing the model to not repeat this behavior and we restart generation.
Section \ref{sec:inference-framework} describes how our system implementation injects information into the model's context when updates arrive from the user and environment. 

\subsection{Speculative Tool Calling}
\label{sec:spec-tool-calling}

Our system aims to launch tool calls as soon as enough information is received from the user or environment to reduce latency. However, there are multiple challenges with launching tool calls based on partial information from the user or environment:

\begin{enumerate}
    \item Even if we know that a tool call is required, we may not yet have received all the information from the environment required to call the tool (e.g. we know that we need to send a message to a given contact, but we are still waiting on an earlier tool call to retrieve the contact information).
    \item It is possible that additional information arrives from the user (or from tool observations) which \emph{invalidates} a tool call which was generated based on partial information, and we need to modify or remove the tool call.
    \item Additionally, not all tool calls can be launched speculatively, as some tool calls modify the system state irreversibly.
\end{enumerate}

In our work, we aim to address these challenges to enable asynchronous reasoning and acting with respect to the user/environment while ensuring correct execution of tool calls.
Our approach builds on the approach from LLMCompiler~\citep{kim2024llm} by managing tool calls as a Directed Acyclic Graph (DAG). 
The existing LLMCompiler method addresses challenge 1) by allowing for emitting tool calls which depend on previous (not yet executed) tool calls. 

To address challenges 2) and 3), we propose \textit{Speculative Tool Calling}, which enables launching tool calls early and later refining tool calls based on updated information from the user or environment.
This approach is visualized in Figure \ref{fig:execution-trace}.
To deal with the challenge that particular ``write'' tools cannot be executed until we are certain that we have received all of the required information (since they irreversibly modify system state), we manually classify tools as either safe or unsafe to execute speculatively based on whether the tool has any side effects (i.e., tools that only read information are marked as safe, whereas tools that modify system state are marked as unsafe), and then hold up execution for unsafe tools until a final confirmation signal is received.
This approach is inspired by how speculative read and write operations are handled in computer hardware memory systems, where write operations are buffered but not allowed to execute until the instruction is committed~\citep{yeager2002mips}. 
Note that this approach still allows for overlapping reasoning and selecting the action while waiting for additional information, so it still reduces latency for tools which are not safe to execute speculatively.

To allow the model to edit tool calls which have already been produced but have not yet been executed, we allow the model to either \textit{modify} or \textit{remove} existing tool calls that have already been issued.
To \textit{modify} a tool call, the model needs to generate a tool call with the same ID as the one that it wants to overwrite (see Sec. \ref{sec:asynchronous-io}).
To \textit{remove} a tool call, the model instead generates a special tool call ``$REMOVE$ $ID.$'', which cancels the tool call with the provided ID.
For tool calls which have executed speculatively and which are later invalidated, if the observation has not yet been inserted into the model's context, we discard it.
When discarding observations, we inject a cancellation message encapsulated in $<cancel>$ / $</cancel>$ tags to inform the model not to expect information to be inserted as
``$<information> ID. ...</information>$''.
If there were any tool calls which depended on the output of the cancelled tool call, these are also cancelled.
For tool calls which are not safe to execute speculatively, if these are modified or removed, the invalid tool call is never executed.
Additionally, to help the model understand which tool calls are active, we provide plan hints each time it produces a tool call or receives a query update (providing the current active tool calls in 
``$<current\_plan>$'' / ``$</current\_plan>$'' tags).

For tools that are not safe to execute speculatively, we allow them to begin execution at the \textit{commit} point.
The \textit{commit} point occurs when the following conditions are met: i) the final query update has been received, and ii) the model either generates a tool call ID greater than any previously generated (implying that any edits to the existing tool calls is complete) or generates a pause action (indicating that there are no other modifications required to the existing plan).
At this point, all tool calls that were held up because they were unsafe to execute speculatively begin execution once all of their dependencies are satisfied.

\subsection{Inference Framework}
\label{sec:inference-framework}

\label{sec:training}

\begin{figure*}[t]
  \centering
  \includegraphics[width=0.85\textwidth]{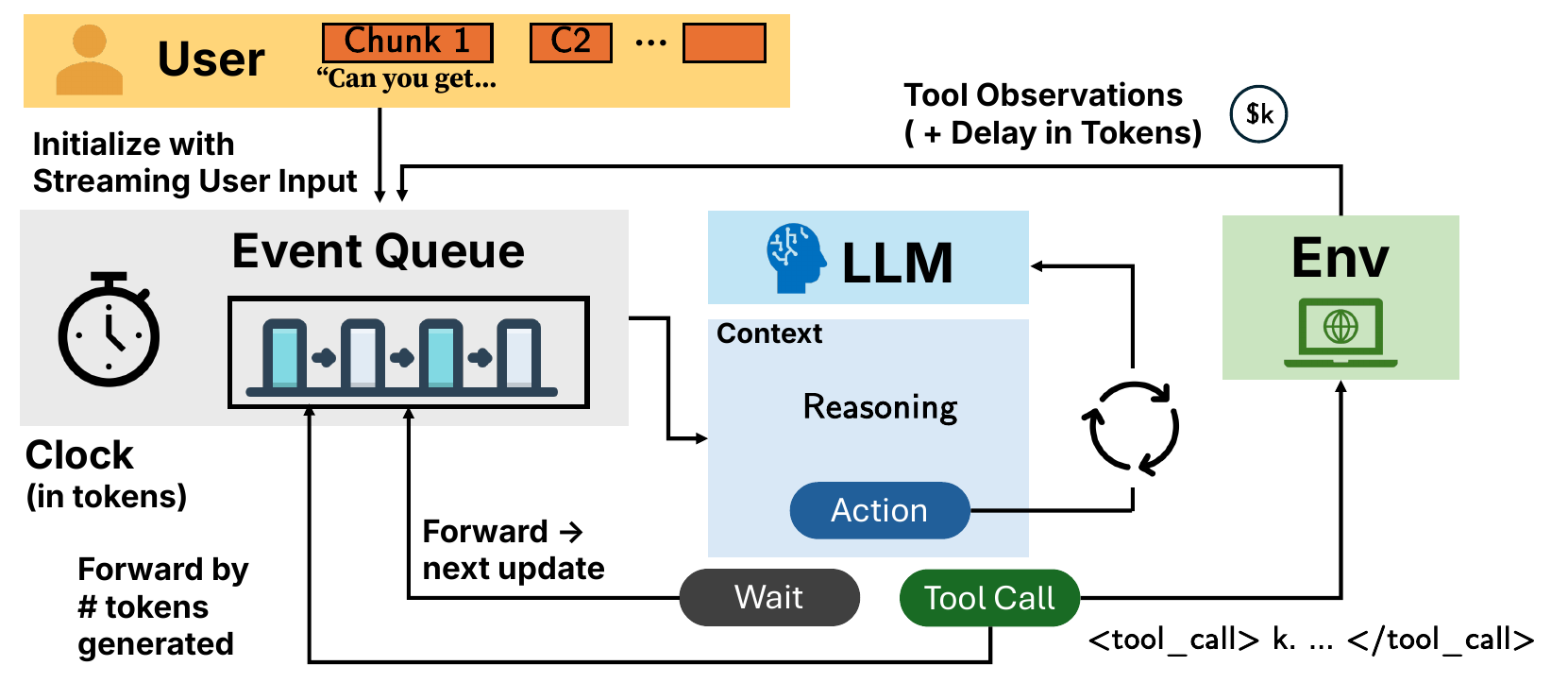}
  \caption{
  \textbf{Training Infrastructure.} A diagram showing how our training infrastructure works to adapt the model to reason and act while waiting on delayed inputs from the user and environment.
  We use an environment which runs on a ``clock'' (with time kept in terms of number of tokens to generate before the model would receive the update). 
  Our approach maintains a central update queue which buffers events until they are ready and then interrupts the model and injects the information into the model's context.
  The model is allowed to reason and act continuously while waiting for additional updates.
  }
  \label{fig:train-system}
\end{figure*}

We present a system implementation to enable the model to reason and act asynchronously with respect to the user and environment, as well as tracking information to manage speculative tool execution.
As described in Figure \ref{fig:system-architecture}, our system is decoupled into a core agentic loop and a task execution manager.
The core agentic loop uses an event-driven framework as in~\citep{rllm2025pepper}. 
At each timestep, the model is allowed to reason and act iteratively based on the currently available information, and it can also wait if no more work can be done until additional information is provided.
When additional information is provided by the user or returned from tool execution, this information can be immediately injected as a new event into a global event queue, and it can then be injected into the model's context.
The core LLM loop uses interruptible streaming with vLLM ~\citep{kwon2023efficient} so that interrupts can be injected immediately as they return.

As in prior work \citep{kim2024llm}, our system leverages a tool execution task manager to launch tool calls as they are ready. 
The manager supports dependency tracking as in \citep{kim2024llm}, which enables launching tool calls which depend on previous tool calls that have not yet finished execution, and which also allows tool calls to execute in parallel if they do not depend on one another.
Similar to \citep{gim2024asynchronous}, as tool calls finish execution, the results can be injected immediately into the context of the model.
Different from these prior efforts, our task manager supports iteratively adding new tasks to the task manager, as well as correcting or removing tool calls from the task queue if later user/environment updates invalidate the earlier tool calls.
Additionally, our task manager also tracks whether tools are safe to execute speculatively, and holds up unsafe tools until the final commit signal is provided.

\section{Training Asynchronous Agents}

\subsection{Training System}
\label{sec:training-details}

To train \textit{Speculative Interaction Agents},
we implement a clock-based training setup (outlined in Figure \ref{fig:train-system}).
This training setup is designed to mimic the inference-time behavior of our system.
In this system, the query updates are initially set at specific time intervals (and the times are converted to numbers of tokens based on an assumed tokens per second rate for decoding) and put in the event queue.
At each step, the model is allowed to generate up to the max number of tokens before the next event is ready to trigger (and we roll the clock forward by the number of tokens that the model actually generated).
If the model doesn't have any work to do until the next update and it decides to wait for an update, we would then roll the clock forward to the time that the next event becomes ready.
When an event is ready, it is inserted into the model's context (and it interrupts the model if it is actively generating).
When the model executes a tool call, we profile the time for the tool call to execute and convert this to the corresponding delay in number of tokens, and then add the tool observation to the global event queue.
This setup simulates the inference-time behavior, where both user inputs and tool calls arrive asynchronously and are inserted into the model's context.
This clock-based setup is used for generating data for SFT.
To simulate streaming, we assume a particular tokens-per-second rate to convert tool delays and user query update delays into numbers of tokens.
Appendix \ref{sec:appendix_experimental_details} describes these details for our evaluation.

\subsection{Data Generation}
\label{sec:data-generation}

To generate data for SFT, we need to simulate streaming interaction between the agent and the environment. 
We first segment the user query into streaming inputs.
To generate arrival timestamps for each segment, we use a text-to-speech (TTS) model to synthesize voice for the segments, and then apply forced alignment using the ground-truth text to get word timestamps.
We use the Kokoro-82M TTS model for voice synthesis, and we use WhisperX to get word alignments ~\citep{kokoro82m,bain2022whisperx}. 
For any sample where forced alignment fails, we filter this out.

Next, we take the ground-truth tool calls from the sample and align them with the earliest update where the sample could have been called.
Since the information required to formulate tool arguments may be semantic or implicit/based on default parameters, we adopt a consistency-based method for identifying the earliest timestep where a function can be called with all the correct arguments.
For each stream update, we use a strong model (Qwen-2.5-32B-Instruct; \citep{qwen2.5}) to generate the tool calls that can be called based on the query so far.
We then work backwards from the final update and identify the earliest query update where each of the ground-truth tool calls could have been called.
Additional details for the data synthesis process are provided in Appendix \ref{sec:appendix_data_generation}.

After obtaining the aligned tool calls, we use the tool calls along with the training gym (outlined in Figure \ref{fig:train-system}) to synthesize trajectories.
At each step, we present the strong model (Qwen-2.5-32B-Instruct) with the trajectory up to the current step, as well as the target action for that step, and we ask the model to synthesize reasoning for the model to arrive at that action. 
The target action for the step is the next tool call that is ready to execute (or to wait if no tool call is ready).
As described in Section \ref{sec:training-details}, our training system simulates streaming behavior and injects interrupts based on the timings for user and environment information, which are converted into numbers of tokens using an assumed tokens-per-second rate.
To synthesize trajectories for the non-streaming baseline, we similarly provide the strong model with the target action at each step and ask the model to generate the reasoning to arrive at that action.

We also provide examples in the training data that show how tool calls can be removed or modified based on updated information in the query.
For TinyAgent in particular, we need to support edits to existing tool calls since it contains sensitive tool calls that need to be able to be modified before being executed if we want to be able to issue them speculatively. 
To adapt the model for this behavior, we synthesize a subset of trajectories with erroneous early tool calls that are later corrected to demonstrate how the model should correct mistakes.
Details for how we synthesize these examples with error correction behavior are explained in Appendix \ref{sec:appendix_data_generation}.

\begin{table*}[!htbp]
\caption{
Evaluation accuracy for the OpenAI Realtime API (openai-realtime-1.5 model) with our Speculative Interaction Agent methodology as well as with the reason-and-act baseline.
We report accuracy on each task as well as latency in seconds (with speedups in parentheses), using 500 samples from each dataset for evaluation due to API cost constraints.
}\label{tab:realtime}
\centering
\scriptsize{
\setlength{\tabcolsep}{5.5pt}{
\begin{tabular}{c|cc|cc}
        \toprule
\multirow{2}{*}{\textbf{Config}} &
\multicolumn{2}{c|}{\textbf{HotpotQA}} & \multicolumn{2}{c}{\textbf{TinyAgent}} 
\\ 
& \textbf{Accuracy} & \textbf{Latency (Speedup)} 
& \textbf{Accuracy} & \textbf{Latency (Speedup)}  \\ 
\midrule
Baseline & 71.6 & 4.5  &  54.9 & 7.6 \\
\hc Speculative Interaction & 71.0 &  3.6 (\textbf{1.3$\times$}) & 53.2 & 4.4 (\textbf{1.7$\times$})  \\
\bottomrule
\end{tabular}
}
}
\end{table*}


\begin{table*}[!htbp]
\caption{
Evaluation accuracy for the Qwen2.5-3B-Instruct and Llama-3.2-3B-Instruct models.
We report results with our Speculative Interaction Agent methodology as well as the reason-and-act baseline, trained using SFT (as well as baseline results without training).
We also report latency in seconds (as well as speedups in parentheses), benchmarked on a single A100 GPU (averaged across 100 samples from the evaluation set). Note that we only report latency results for the baseline with normal SFT and for our methodology with Speculative Interaction SFT (SI-SFT), as the other configurations struggle to follow the desired behavior (e.g. not executing any tool calls before responding) which makes their latency not directly comparable.
}
\label{tab:trained}
\vspace{1mm}
\scriptsize
\centering
\renewcommand{\arraystretch}{1.2}
\setlength{\tabcolsep}{6pt}
\begin{tabular}{c|cc|cc|cc|cc}
\toprule
\multirow{3}{*}{\textbf{Config}} &
\multicolumn{4}{c|}{\textbf{Qwen2.5-3B}} &
\multicolumn{4}{c}{\textbf{Llama-3.2-3B}} \\
\cmidrule{2-9}
& \multicolumn{2}{c|}{\textbf{HotpotQA}} & \multicolumn{2}{c|}{\textbf{TinyAgent}} & \multicolumn{2}{c|}{\textbf{HotpotQA}} & \multicolumn{2}{c}{\textbf{TinyAgent}} \\ 
& \textbf{Acc.} & \textbf{Lat. } & \textbf{Acc.} & \textbf{Lat. } & \textbf{Acc.} & \textbf{Lat. } & \textbf{Acc.} & \textbf{Lat. }  \\ 
\midrule\midrule
Baseline (no SFT) & 17.2 & - & 3.2 & - & 15.1 & - & 1.5 & - \\
Baseline (normal SFT)     & 68.6 & 2.7 & 65.6 & 4.1 & 70.4 & 2.3 & 66.8 & 5.0 \\
\midrule
Speculative Interaction (no SFT)     & 20.1 & - & 2.3 & - & 13.9 & - & 2.1 & - \\
\hc Speculative Interaction (normal SFT)     & 33.4 & - & 14.3 & - & 23.6 & - & 10.8 & - \\
\hd Speculative Interaction (SI-SFT)     & 67.5 & 1.2 (\textbf{2.2$\times$})
 & 62.1 & 2.5 (\textbf{1.6$\times$}) & 68.7 & 1.1 (\textbf{2.1$\times$}) & 65.2 & 2.5 (\textbf{2.0$\times$})\\
\bottomrule
\end{tabular}
\end{table*}

\section{Results}

\subsection{Evaluation}

We evaluate our method across one closed-source cloud API (the OpenAI Realtime API with openai-realtime-1.5 ~\citep{openai_realtime_api}) and two open-source language models (Qwen2.5-3B-Instruct~\citep{qwen2.5} and Llama-3.2-3B-Instruct~\citep{grattafiori2024llama}). 
For the closed-source model, we provide multiple in-context examples to guide model behavior.
We evaluate on HotpotQA as a comparison task requiring multiple search tool calls~\citep{yang2018hotpotqa}.
For more complex agentic tasks, we include multi-step evaluation with TinyAgent~\citep{erdogan2024tinyagentfunctioncallingedge}, which evaluates multi-step personal assistant-style workflows for a local agent on a laptop. 
Across both tasks, we report both accuracy (determined based on state-based correctness evaluation) and latency in seconds.
For latency evaluation, we simulate time delays using arrival timestamps obtained from forced alignment applied to the TTS outputs (as outlined in Section \ref{sec:data-generation}).

\subsection{Experimental Details}
\label{sec:experimental-details}

We evaluate our method by simulating streaming updates from speech-to-text.
We assume 6-word increments for TinyAgent and 3-word increments for HotpotQA due to the shorter average query length.
For latency evaluation, we use a fake streaming speech-to-text frontend to simulate the arrival of partial user query updates as the user is speaking.
We leverage the timings obtained through TTS alignment (see Sec. \ref{sec:data-generation}) to simulate when speech would arrive from the user.
For HotpotQA, we use the real tool latencies when measuring latency.
For TinyAgent, we are using a simulated environment for verifying state correctness (described in detail in Appendix \ref{sec:appendix_experimental_details}); we simulate the delay for each tool call as a random uniform delay between 0.5 and 1 seconds.
We provide additional experimental details (including training details) in Appendix \ref{sec:appendix_experimental_details}.

\subsection{Main Results}

The main evaluation results for the OpenAI realtime API are provided in Table \ref{tab:realtime}.
We present results for both the synchronous reason-and-act baseline, as well as the Speculative Interaction Agent methodology.
We observe speedups of 1.3-1.7$\times$ with minor accuracy loss.
Table \ref{tab:trained} also provides evaluation for open-source models on HotpotQA and TinyAgent.
We present results for the synchronous reason-and-act baseline, as well as for the Speculative Interaction Agent method.
For both methods, we report results with SFT, and for the baseline we also report results without training.
Additionally, we observe \textbf{1.6-2.2}$\times$ average speedups across both models and tasks relative to the strong SFT baseline, and we attain close to the accuracy of the non-streaming SFT runs.

\subsection{Evaluation Dataset of Naturalistic Utterances}

In our evaluation, we aim to capture realistic challenges with real-time human-AI interactions that are not captured by existing benchmarks. 
For example, realistic, \emph{spontaneous} user utterances feature: (1) filler words and hesitation markers (e.g. 'um', 'yeah'), (2) online clarification of ambiguities (e.g. ``which I mean by that is I want..``), and (3) modification or corrections of requests (e.g. ``wait, actually could you instead ...``). 
To address this, we present a challenging human-produced naturalistic evaluation dataset that complements existing benchmarks.
This dataset includes 177 samples based on TinyAgent and its tool set.
Appendix \ref{sec:appendix_human_dataset} outlines details for our dataset construction, as well as results when evaluating our method on challenging naturalistic samples.
\section{Conclusion}

There is a growing demand for real-time interaction with agentic AI technologies to align their behavior with user demands; however, complex agentic workflows typically add several seconds of latency due to the time for the model to reason for each action as well as long tool latencies.
In this work, we present an algorithm and system design which enables real-time interaction with agentic systems with low latency.
We propose \textit{Asynchronous I/O}, where we decouple the agent's thinking and acting from delays with information arriving from the user and environment.
Our \textit{Speculative Tool Calling} approach allows the model to launch tool calls and later correct any mistakes, and ensures that sensitive tools do not execute until we have ensured their arguments are correct based on the full user input.
We present an event-driven inference framework which contains a task execution manager as well as a core agent loop which supports interrupting agent execution when new information arrives.
We also propose a clock-based training methodology for generating synthetic data for SFT.
Altogether, our \textit{Speculative Interaction Agent} methodology provides up to 2.2$\times$ speedup relative to the reason and act baseline.

\section{Acknowledgements}

We acknowledge gracious support from the Furiosa AI, Intel, Apple, NVIDIA, Macronix, Panasonic, and Mozilla teams. 
Furthermore, we appreciate support from Google Cloud, the Google TRC team, and specifically Jonathan Caton, Divvy Thakkar, and Prof. David Patterson.
Prof. Keutzer’s lab is also sponsored by funding through BDD and BAIR. 
We also acknowledge support by the Director, Office of Science, Office of Advanced Scientific Computing Research, of the U.S. Department of Energy under
Contract No. DE-AC02-05CH11231. Michael W. Mahoney acknowledges
DARPA, NSF, the DOE Competitive Portfolios grant, and
the DOE SciGPT grant. Our conclusions do not necessarily
reflect the position or the policy of our sponsors, and no
official endorsement should be inferred.

\bibliographystyle{colm2026_conference}
\bibliography{sample-base}

@String{Computer = "{IEEE} Computer" }

@inproceedings{kim2024llm,
  title={An llm compiler for parallel function calling},
  author={Kim, Sehoon and Moon, Suhong and Tabrizi, Ryan and Lee, Nicholas and Mahoney, Michael W and Keutzer, Kurt and Gholami, Amir},
  booktitle={Forty-first International Conference on Machine Learning},
  year={2024}
}

@article{arora2025stream,
  title={Stream rag: Instant and accurate spoken dialogue systems with streaming tool usage},
  author={Arora, Siddhant and Khan, Haidar and Sun, Kai and Dong, Xin Luna and Choudhary, Sajal and Moon, Seungwhan and Zhang, Xinyuan and Sagar, Adithya and Appini, Surya Teja and Patnaik, Kaushik and others},
  journal={arXiv preprint arXiv:2510.02044},
  year={2025}
}

@article{chiang2025shanks,
  title={SHANKS: Simultaneous Hearing and Thinking for Spoken Language Models},
  author={Chiang, Cheng-Han and Wang, Xiaofei and Li, Linjie and Lin, Chung-Ching and Lin, Kevin and Liu, Shujie and Wang, Zhendong and Yang, Zhengyuan and Lee, Hung-yi and Wang, Lijuan},
  journal={arXiv preprint arXiv:2510.06917},
  year={2025}
}

@article{chiang2025stitch,
  title={Stitch: Simultaneous thinking and talking with chunked reasoning for spoken language models},
  author={Chiang, Cheng-Han and Wang, Xiaofei and Li, Linjie and Lin, Chung-Ching and Lin, Kevin and Liu, Shujie and Wang, Zhendong and Yang, Zhengyuan and Lee, Hung-yi and Wang, Lijuan},
  journal={arXiv preprint arXiv:2507.15375},
  year={2025}
}

@article{yakushev2025asynchronous,
  title={Asynchronous Reasoning: Training-Free Interactive Thinking LLMs},
  author={Yakushev, George and Babina, Nataliia and Dastgerdi, Masoud Vahid and Zhdanovskiy, Vyacheslav and Shutova, Alina and Kuznedelev, Denis},
  journal={arXiv preprint arXiv:2512.10931},
  year={2025}
}

@article{gim2024asynchronous,
  title={Asynchronous LLM Function Calling},
  author={Gim, In and Lee, Seung-seob and Zhong, Lin},
  journal={arXiv preprint arXiv:2412.07017},
  year={2024}
}

@article{wang2025staircase,
  title={Staircase Streaming for Low-Latency Multi-Agent Inference},
  author={Wang, Junlin and Wang, Jue and Athiwaratkun, Ben and Dhingra, Bhuwan and Zhang, Ce and Zou, James and others},
  journal={arXiv preprint arXiv:2510.05059},
  year={2025}
}

@article{rllm2025pepper,
  title = {Pepper: A Real‑Time, Event‑Driven Architecture for Proactive Agentic Systems},
  author = {Wu, Tianhao and Tan, Sijun and rLLM Team},
  journal = {rLLM Blog},
  year = {2025},
  month = {October},
  url = "https://rllm-project.com/post.html?post=pepper.md"
}

@article{shih2025can,
  title={Can Speech LLMs Think while Listening?},
  author={Shih, Yi-Jen and Raj, Desh and Wu, Chunyang and Zhou, Wei and Bong, SK and Gaur, Yashesh and Mahadeokar, Jay and Kalinli, Ozlem and Seltzer, Mike},
  journal={arXiv preprint arXiv:2510.07497},
  year={2025}
}

@article{li2025predgen,
  title={PredGen: Accelerated Inference of Large Language Models through Input-Time Speculation for Real-Time Speech Interaction},
  author={Li, Shufan and Grover, Aditya},
  journal={arXiv preprint arXiv:2506.15556},
  year={2025}
}

@article{thinkingmachines2026interactionmodels,
  author = {Thinking Machines Lab},
  title = {Interaction Models: A Scalable Approach to Human-AI Collaboration},
  journal = {Thinking Machines Lab: Connectionism},
  year = {2026},
  month = {May},
  note = {https://thinkingmachines.ai/blog/interaction-models/},
  doi = {10.64434/tml.20260511},
}

@article{ok2025speculative,
  title={Speculative End-Turn Detector for Efficient Speech Chatbot Assistant},
  author={Ok, Hyunjong and Yoo, Suho and Lee, Jaeho},
  journal={arXiv preprint arXiv:2503.23439},
  year={2025}
}

@article{acikgoz2025speakrl,
  title={SpeakRL: Synergizing Reasoning, Speaking, and Acting in Language Models with Reinforcement Learning},
  author={Acikgoz, Emre Can and Oh, Jinoh and Hao, Jie and Jeon, Joo Hyuk and Ji, Heng and Hakkani-T{\"u}r, Dilek and Tur, Gokhan and Li, Xiang and Ma, Chengyuan and Fan, Xing},
  journal={arXiv preprint arXiv:2512.13159},
  year={2025}
}

@article{yeager2002mips,
  title={The MIPS R10000 superscalar microprocessor},
  author={Yeager, Kenneth C},
  journal={IEEE micro},
  volume={16},
  number={2},
  pages={28--41},
  year={2002},
  publisher={IEEE}
}

@inproceedings{kwon2023efficient,
  title={Efficient memory management for large language model serving with pagedattention},
  author={Kwon, Woosuk and Li, Zhuohan and Zhuang, Siyuan and Sheng, Ying and Zheng, Lianmin and Yu, Cody Hao and Gonzalez, Joseph and Zhang, Hao and Stoica, Ion},
  booktitle={Proceedings of the 29th symposium on operating systems principles},
  pages={611--626},
  year={2023}
}

@inproceedings{miller1968response,
  title={Response time in man-computer conversational transactions},
  author={Miller, Robert B},
  booktitle={Proceedings of the December 9-11, 1968, fall joint computer conference, part I},
  pages={267--277},
  year={1968}
}

@misc{kokoro82m,
  author       = {hexgrad},
  title        = {Kokoro-82M},
  year         = {2025},
  publisher    = {Hugging Face},
  howpublished = {\url{https://huggingface.co/hexgrad/Kokoro-82M}},
  doi          = {10.57967/hf/4329}
}

@article{bain2022whisperx,
  title={WhisperX: Time-Accurate Speech Transcription of Long-Form Audio},
  author={Bain, Max and Huh, Jaesung and Han, Tengda and Zisserman, Andrew},
  journal={INTERSPEECH 2023},
  year={2023}
}

@misc{qwen2.5,
    title = {Qwen2.5: A Party of Foundation Models},
    url = {https://qwenlm.github.io/blog/qwen2.5/},
    author = {Qwen Team},
    month = {September},
    year = {2024}
}

@misc{erdogan2024tinyagentfunctioncallingedge,
      title={TinyAgent: Function Calling at the Edge},
      author={Lutfi Eren Erdogan and Nicholas Lee and Siddharth Jha and Sehoon Kim and Ryan Tabrizi and Suhong Moon and Coleman Hooper and Gopala Anumanchipalli and Kurt Keutzer and Amir Gholami},
      year={2024},
      eprint={2409.00608},
      archivePrefix={arXiv},
      primaryClass={cs.CL},
      url={https://arxiv.org/abs/2409.00608},
}

@article{grattafiori2024llama,
  title={The llama 3 herd of models},
  author={Grattafiori, Aaron and Dubey, Abhimanyu and Jauhri, Abhinav and Pandey, Abhinav and Kadian, Abhishek and Al-Dahle, Ahmad and Letman, Aiesha and Mathur, Akhil and Schelten, Alan and Vaughan, Alex and others},
  journal={arXiv preprint arXiv:2407.21783},
  year={2024}
}

@inproceedings{yang2018hotpotqa,
  title={{HotpotQA}: A Dataset for Diverse, Explainable Multi-hop Question Answering},
  author={Yang, Zhilin and Qi, Peng and Zhang, Saizheng and Bengio, Yoshua and Cohen, William W. and Salakhutdinov, Ruslan and Manning, Christopher D.},
  booktitle={Conference on Empirical Methods in Natural Language Processing ({EMNLP})},
  year={2018}
}

@misc{openai_realtime_api,
  author = {OpenAI},
  title = {OpenAI Realtime {API}},
  year = {2024},
  howpublished = {\url{https://platform.openai.com/docs/guides/realtime}},
  note = {Accessed: [Insert Date Here]}
}

@article{wei2022chain,
  title={Chain-of-thought prompting elicits reasoning in large language models},
  author={Wei, Jason and Wang, Xuezhi and Schuurmans, Dale and Bosma, Maarten and Xia, Fei and Chi, Ed and Le, Quoc V and Zhou, Denny and others},
  journal={Advances in neural information processing systems},
  volume={35},
  pages={24824--24837},
  year={2022}
}

@article{guo2025deepseek,
  title={Deepseek-r1: Incentivizing reasoning capability in llms via reinforcement learning},
  author={Guo, Daya and Yang, Dejian and Zhang, Haowei and Song, Junxiao and Zhang, Ruoyu and Xu, Runxin and Zhu, Qihao and Ma, Shirong and Wang, Peiyi and Bi, Xiao and others},
  journal={arXiv preprint arXiv:2501.12948},
  year={2025}
}

@misc{qwq32b,
    title = {QwQ-32B: Embracing the Power of Reinforcement Learning},
    url = {https://qwenlm.github.io/blog/qwq-32b/},
    author = {Qwen},
    month = {March},
    year = {2025}
}

@inproceedings{yao2022react,
  title={React: Synergizing reasoning and acting in language models},
  author={Yao, Shunyu and Zhao, Jeffrey and Yu, Dian and Du, Nan and Shafran, Izhak and Narasimhan, Karthik R and Cao, Yuan},
  booktitle={The eleventh international conference on learning representations},
  year={2022}
}

@article{patil2024gorilla,
  title={Gorilla: Large language model connected with massive apis},
  author={Patil, Shishir G and Zhang, Tianjun and Wang, Xin and Gonzalez, Joseph E},
  journal={Advances in Neural Information Processing Systems},
  volume={37},
  pages={126544--126565},
  year={2024}
}

@article{schick2023toolformer,
  title={Toolformer: Language models can teach themselves to use tools},
  author={Schick, Timo and Dwivedi-Yu, Jane and Dess{\`\i}, Roberto and Raileanu, Roberta and Lomeli, Maria and Hambro, Eric and Zettlemoyer, Luke and Cancedda, Nicola and Scialom, Thomas},
  journal={Advances in Neural Information Processing Systems},
  volume={36},
  pages={68539--68551},
  year={2023}
}

@misc{vllm2026streaming,
  title   = {Streaming Requests \& Realtime {API} in {vLLM}},
  author  = {{Meta, Mistral AI, and the vLLM team}},
  year    = {2026},
  month   = jan,
  url     = {https://blog.vllm.ai/2026/01/31/streaming-realtime.html},
  note    = {vLLM Blog},
}

@online{GoogleLiveAPI2025,
  author       = {{Google Cloud}},
  title        = {Multimodal Live API Reference},
  year         = {2025},
  url          = {https://cloud.google.com/vertex-ai/generative-ai/docs/model-reference/multimodal-live},
  urldate      = {2026-02-14},
  organization = {Google LLC},
  note         = {Vertex AI Documentation}
}

@misc{jeffries2024moonshine,
  title={Moonshine: Speech Recognition for Live Transcription and Voice Commands}, 
  author={Nat Jeffries and Evan King and Manjunath Kudlur and Guy Nicholson and James Wang and Pete Warden},
  year={2024},
  eprint={2410.15608},
  archivePrefix={arXiv},
  primaryClass={cs.SD},
  url={https://arxiv.org/abs/2410.15608}
}

@inproceedings{zheng2024llamafactory,
  title={LlamaFactory: Unified Efficient Fine-Tuning of 100+ Language Models},
  author={Yaowei Zheng and Richong Zhang and Junhao Zhang and Yanhan Ye and Zheyan Luo and Zhangchi Feng and Yongqiang Ma},
  booktitle={Proceedings of the 62nd Annual Meeting of the Association for Computational Linguistics (Volume 3: System Demonstrations)},
  address={Bangkok, Thailand},
  publisher={Association for Computational Linguistics},
  year={2024},
  url={http://arxiv.org/abs/2403.13372}
}

\appendix
\newpage
\section{Extended Related Work}

\label{sec:appendix_related_work}

Here, we provide an extended discussion of related work on efficient language and voice agents.

\subsection{Efficient LLM Agents}

Several prior works have explored how to execute agentic workflows more efficiently, including parallelizing tool calls where possible and performing work while waiting on asynchronous tool execution responses.
LLMCompiler \citep{kim2024llm} proposes a separation of concerns between tool call planning and execution, and allows tool calls to be executed in parallel when they do not depend on one another.
\citep{gim2024asynchronous} proposed Asynchronous LLM Function Calling, where the model can keep working on subsequent tool calls while waiting for earlier tool calls to execute.

There has also been work on designing more proactive and interactive LLM agents.
Pepper \citep{rllm2025pepper} proposed an event-driven architecture for proactive LLM assistants.
\citep{acikgoz2025speakrl} proposed an RL-based method for training agents to ask for clarifications for interactive human-agent collaboration. 
\citep{yakushev2025asynchronous} proposed a training-free method to simultaneously run separate thinker and talker threads, so that the model could respond to the user while running a background reasoning process.
\citep{wang2025staircase} proposed a staircase streaming method for multi-agent systems, where agents could begin acting before receiving the full communication from other agents.
Our approach builds on this prior work to propose a general framework for reasoning and acting asynchronously with respect to both user inputs and the environment.

Compared with prior work, our work builds on the asynchronous tool calling method from \citep{gim2024asynchronous} and generalizes this to handle both asynchronous user updates as well as asynchronous responses from the environment.
Our work also builds on the decoupled planning/execution from LLMCompiler \citep{kim2024llm}, which allows for tool calls to be executed in parallel as soon as their operands are ready.
We extend this to allow for issuing tool calls iteratively as we obtain sufficient information, and also propose \textit{Speculative Tool Calling} (Sec. \ref{sec:spec-tool-calling}) to hold up sensitive tool calls until they are confirmed and to allow for modifying/removing tool calls that have been issued but not yet finished execution.

\subsection{Efficient Voice Agents}

There has also been prior work which has explored methods to accelerate voice agents by exploiting the streaming nature of speech.
StreamRAG \citep{arora2025stream} proposes launching RAG calls partway through receiving the streaming user input, thereby reducing response latency.
SHANKS \citep{chiang2025shanks} processes the streaming input speech in fixed chunks and interleaves unspoken reasoning (as well as tool calls) between chunks while waiting for further information from the user, whereas STITCH \citep{chiang2025stitch} interleaved text-to-speech with unspoken reasoning.
\citep{shih2025can} also demonstrated how thinking can be overlapped with listening to streaming user inputs. 

Additionally, previous works have explored methods for speculating on the remaining user response when processing streaming speech inputs to reduce latency.
PredGen \citep{li2025predgen} started generating the user response by speculating on the end of the user input, thereby reducing latency. 
\citep{ok2025speculative} also proposed speculative end-of-turn detection to reduce latency by beginning processing the user input early.

Compared with the most similar existing work (SHANKS \citep{chiang2025shanks}), we pursue a similar approach where we interleave reasoning and acting with streaming input speech. 
The key differences with our method are 1) we support asynchronous tool calling as well as asynchronous user inputs, 2) we propose \textit{Speculative Tool Calling} (Sec. \ref{sec:spec-tool-calling}) to handle sensitive tool calls and cases where the tool call is issued incorrectly with partial information. 

\section{Data Generation}
\label{sec:appendix_data_generation}

\subsection{Additional Data Generation Details}

During the alignment process (where we take the ground-truth tool calls from the sample and align them with the earliest update where the sample could have been called), we apply post-processing to improve alignment and to remove erroneous samples from the training set. 
With our consistency-based method, for each stream update, we use a strong model (Qwen-2.5-32B-Instruct \citep{qwen2.5}) to generate the tool calls that can be called based on the query so far.
We then work backwards from the final update and identify the earliest query update where each of the ground-truth tool calls could have been called.
For both datasets, we use embedding-based string matching to align the ground-truth tool calls with the corresponding tool calls executed by the strong model at each step, and we filter out samples where there is a mismatch between the ground-truth plan and the tool calls generated by the strong model.
Additionally, for HotpotQA, which consists of search calls where the entity name can often be matched in the provided query, we apply an additional post-processing step to adjust the timings in cases where a matching entity name was available earlier or later than the step where the tool call was first correctly issued. 

\subsection{Trajectories with Error Correction Behavior}

To teach the model to call tools speculatively and then fix any mistakes, we need to provide examples in the training data that show how tool calls can be removed or modified based on updated information in the query. 
To adapt the model for this behavior, we synthesize a subset of trajectories with erroneous early tool calls that are later corrected to demonstrate how the model should correct mistakes.
To do this, while aligning ground-tool calls, we identify tool calls which are initially issued incorrectly and which are no longer present in the final plan, and we record when these erroneous tool calls are issued (i.e. after which query update).
We separate these erroneous tool calls into ones which are later modified to a correct tool call versus ones which have to be removed by identifying whether there is a similar tool call in the final ground-truth plan (using embedding-based string matching).
When synthesizing trajectories, we ask the model to generate the reasoning to issue these erroneous actions, and then at a later timestep we ask the model to fix the incorrect action once sufficient information is available to issue the correction.

\section{Extended Experimental Details}
\label{sec:appendix_experimental_details}
For accuracy evaluation, we use the final response or state to evaluate correctness (as if we matched the produced tool calls exactly with the ground truth, this would potentially mark a sample as wrong due to a speculatively launched tool call, even if this was later corrected). 
For HotpotQA, we use the final response after the model has searched for information and compare it with the ground truth answer.
For TinyAgent, since the required response is often to actuate a series of actions instead of just returning information to the user, we use state-based evaluation to verify whether the user's query was satisfied by the response.
We construct a simulated environment for TinyAgent which is initialized with data that read tool calls can access and which stores data whenever we execute a tool call that writes data.
To check correctness, we execute the ground-truth tool calls in a copy of the simulated environment, and compare this with a separate copy of the simulated environment after running evaluation for that sample.

For TinyAgent, we use a simulated setup which assumes a random uniform delay of between 0.5 and 1 seconds for each tool call, whereas for HotpotQA, we use the real tool latencies.
To measure accuracy with streaming for open-source models, we assume a tokens-per-second generation rate randomly sampled between 100 and 200 tokens/second for each sample, and we convert both the streaming update times and the tool call delays to numbers of tokens in order to simulate streaming.
We assume updates are passed to the model in 6-word increments; note that for HotpotQA we lessen this to 3-word increments due to the short average query length.
For the closed-source API, to minimize API costs, we run the accuracy and time-to-first-token evaluation together using the real-world tool latencies and streaming inference delays from the API.

For latency evaluation, we use a fake streaming speech-to-text frontend to simulate the arrival of partial user query updates as the user is speaking.
We leverage the timings obtained through TTS alignment (see Sec. \ref{sec:data-generation}) to simulate when speech would arrive from the user.
For HotpotQA, we use the real tool latencies when measuring latency.
For TinyAgent, since we are using a simulated environment for verifying state correctness, we simulate the delay for each tool call as a random uniform delay between 0.5 and 1 seconds.

For data generation for SFT, we assume the same tokens-per-second rate and tool call delays as are used in the evaluation process (we randomly sampled between 100 and 200 tokens/second for each sample, and we use either real tool call delays or a random uniform delay of between 0.5 and 1 seconds for each tool call, dependending on the dataset).
For 10\% of samples, we include incorrectly generated tool calls which the model has to either remove or modify once additional information is available.
We use LLaMA-Factory for finetuning experiments \citep{zheng2024llamafactory}.
For streaming experiments, we apply masking to ensure that when the model is interrupted, no loss is applied to the final generated token from the model (as it should not learn to stop at that position).
We train for 1 epoch using a batch size of 16, learning rate of 3e-5, warmup of 0.05 using a cosine LR scheduler, and a weight decay of 0.05.
\section{Human Instructions Dataset}
\label{sec:appendix_human_dataset}
\subsection{Dataset Details}

Our Human Instructions Dataset consists of real-world voice-assistant interactions paired with LLM-generated execution plans which have gone under LLM-as-a-judge and subsequent human review. Each dataset entry contains: (1) a verbatim transcript of a spoken user query and (2) a structured plan consisting of a sequence of tool/function calls intended to fulfill the user’s request. 
We aim to use these samples to evaluate our streaming methodology on challenging real-world samples with voice transcription challenges or where the user's intent drifts. 

\subsection{Data Collection Process Details}

Our team of 24 project contributors produced evaluation samples for Human Instructions. 
In each session, a contributor was provided with a description of the available tools and their functionalities, based on which they generated a diverse set of tasks that could be accomplished using these tools.
The generated tasks range from extremely simple requests that required a single tool invocation to more complex scenarios involving multi-step workflows and mid-utterance intent changes. 
A total of 278 sessions were produced initially; after post-processing, we retain 177 sessions that are used as final evaluation data.

All interactions were recorded as raw audio. In addition to the transcript text, we retained timestamp annotations for each utterance, enabling reconstruction of the temporal structure of the original speech.
Since questions were derived from real voice transcripts, they naturally contain speech disfluencies including filler words such as "uh" or "um", pauses (``...''), repetition, and imperfect grammar. These features are preserved to maintain transcript fidelity. Transcripts were generated using Deepgram \texttt{nova-2-general} and are not normalized to preserve realistic downstream evaluation conditions. 

\subsection{Post-Processing}

\paragraph{Initial Transcript Set.}
We begin with a corpus of real voice-assistant transcripts. Labels or annotations are not present at this stage.

\paragraph{Generation of Candidate Plans.}
To construct transcript--plan pairs, we generate execution plans using GPT-5.2 in a zero-label evaluation setting. The model is prompted to produce structured tool-call sequences that fulfill each transcript’s request. This produces an initial dataset of 278 entries consisting of transcripts paired with model-generated plans.

\paragraph{LLM-as-Judge Annotation.}
To assign quality labels, we use a structured judging prompt (shown in section A.4) that defines the classification task and evaluation criteria. The prompt is augmented with few-shot examples to calibrate model behavior. Notably, each entry is assigned to exactly one of the following four mutually exclusive categories:

\begin{description}

\item[\textbf{Perfect.}] The question is coherent and the generated plan correctly fulfills the request using the appropriate functions, arguments, and ordering. No modifications are required.

\item[\textbf{Passable.}] The question contains minor issues but the intent remains interpretable, and the generated plan reasonably satisfies the request. The judge must provide minimal one-to-one word substitutions to clarify ambiguities while preserving transcript fidelity. If necessary, corresponding answer fixes must also be provided.

\item[\textbf{LLM\_Wrong.}] The question is understandable but the generated plan is incorrect, incomplete, improperly ordered, or fails to account for user intent (including changes of mind). The judge must provide a corrected plan together with an explanation of the error.

\item[\textbf{Delete.}] The request is either incoherent or fundamentally unachievable with the available tools. In this case, no fixes are produced.

\end{description}

In addition to assigning a category label, judges are required to produce a natural-language explanation justifying their decision. When edits are proposed to either the question or the answer, the judge must separately provide the corrected text and an accompanying explanation.

\paragraph{Question Fix Guidelines.}
Question fixes are strictly limited to one-to-one word substitutions and must result in a coherent request without rewriting, normalization, or stylistic modification of the transcript. This constraint is necessary because we preserve word-level timestamp annotations, which are used to replicate natural speech characteristics in both training and evaluation settings.

\paragraph{Answer Fix Guidelines.}
Answer fixes must produce a corrected execution plan that fully satisfies the user’s request using valid tool calls, correct arguments, and proper ordering. The corrected plan must explicitly include all initial tool calls even when the user changes their mind. Judges may not introduce unsupported tools, omit required calls, or alter the task definition beyond what is necessary to faithfully execute the clarified request.

Each entry is evaluated independently by three judge models:
\begin{itemize}
    \item GPT-5.2 (\texttt{gpt-5.2-2025-12-11})
    \item Gemini 3.1 Pro Preview
    \item Claude Opus 4.6
\end{itemize}

\paragraph{Aggregation via Majority Voting.}
Final labels are assigned using majority vote across the three judge models. The resulting distribution is:

\begin{center}
\begin{tabular}{lr}
\textbf{Label} & \textbf{Count} \\
\hline
delete & 91 \\
llm\_wrong & 114 \\
perfect & 43 \\
passable & 16 \\
three-way split & 14 \\
\end{tabular}
\end{center}

\paragraph{Human Verification.}
After automatic labeling, a thorough human review pass is conducted to validate label correctness, resolve three-way disagreements, and correct any errors. The final data set statistics after human verification are reported below:

\vspace{0.5em}
\begin{center}
\begin{tabular}{lr}
\textbf{Label} & \textbf{Count} \\
\hline
delete & 101 \\
valid & 177
\end{tabular}
\end{center}

\subsection{Evaluation Results}

Table \ref{tab:human-instructions} provides evaluation for the Qwen2.5-3B-Instruct and Llama-3.2-3B-Instruct models on Human Instructions. 
We evaluate the base checkpoints and baseline SFT checkpoints without streaming, and we evaluate our method using the post-SI-SFT model.
We use the SFT checkpoints trained using the TinyAgent training set since the provided tool set is the same.
On Human Instructions, we observe substantially lower accuracy relative to evaluation on TinyAgent, even with the post-SFT models. 
We also observe a larger discrepancy between the normal-SFT baseline and our Speculative Interaction Agent model adapted using SI-SFT.
Additionally, we observe higher latency with the Speculative Interaction Agent method relative to the baseline, as we observe degenerative behavior where the model produces repeated actions.
These results highlight that there is room for further development in model adaptation strategies for processing streaming user inputs with naturalistic speech.

\begin{table*}[!htbp]
\caption{
Evaluation accuracy for the Qwen2.5-3B-Instruct and Llama-3.2-3B-Instruct models on Human Instructions.
We report results with our Speculative Interaction Agent methodology as well as the reason-and-act baseline, trained using SFT (as well as baseline results without training).
We also report latency in seconds, which was benchmarked on a single A100 GPU (averaged across 10 samples from the evaluation set). 
Note that we only report latency results for the baseline with normal SFT and for Speculative Interaction Agent with SI-SFT, as the other configuration struggled to follow the desired behavior (e.g. not executing any tool calls before responding) which makes the latency not directly comparable.
}\label{tab:human-instructions}
\centering
\scriptsize{
\setlength{\tabcolsep}{5.5pt}{
\begin{tabular}{c|cc|cc}
        \toprule
\multirow{2}{*}{\textbf{Config}} &
\multicolumn{2}{c|}{\textbf{Qwen2.5-3B}} & \multicolumn{2}{c}{\textbf{Llama-3.2-3B}} 
\\ 
& \textbf{Accuracy} & \textbf{Latency} 
& \textbf{Accuracy} & \textbf{Latency}  \\ 
\midrule
Baseline (no SFT) & 4.0 & -  &  4.0 & -\\
Baseline (normal SFT) & 22.0 & 9.3  &  19.2 & 5.6 \\
Speculative Interaction (SI-SFT) & 13.6 & 14.3 & 15.3 & 9.3 \\
\bottomrule
\end{tabular}
}
}
\end{table*}

\end{document}